\documentclass{CVIS}

\usepackage{amsmath,bm}
\usepackage{epstopdf}
\begin{document}

\title{Domain Adaptation and Transfer Learning in StochasticNets}

\author{
\begin{tabularx}{\textwidth}{X X}
Mohammad Javad Shafiee & University of Waterloo, ON, Canada \\
Parthipan Siva & Aimetis Corp., ON, Canada \\
Paul Fieguth & University of Waterloo, ON, Canada \\
Alexander Wong & University of Waterloo, ON, Canada
\end{tabularx}
}

\maketitle

\begin{abstract}
\vspace{-0.2 cm}
Transfer learning is a recent field of machine learning research that aims to resolve the challenge of dealing with insufficient training data in the domain of interest.  This is a particular issue with traditional deep neural networks where a large amount of training data is needed.  Recently, StochasticNets was proposed to take advantage of sparse connectivity in order to decrease the number of parameters that needs to be learned, which in turn may relax training data size requirements.   In this paper, we study the efficacy of transfer learning on StochasticNet frameworks. Experimental results show $\sim$7\% improvement on StochasticNet performance when the transfer learning is applied in training step.
\end{abstract}
\vspace{- 0.5 cm}
\section{Introduction}
\vspace{-0.4 cm}
In most machine learning algorithms, it is assumed that the training data and future data must be in the same feature space and follow the same distribution~\cite{Tlearning}.  However, this assumption often does not hold in real-world situations where one wishes to perform classification in one domain but there is only sufficient training data in another domain.  This issue of insufficient training data in the domain of interest is particularly problematic for deep neural networks, which are characterized by a large number of parameters that needs to be learned and as such requires a large amount of training data to learn optimal model parameters.  

In an attempt to reduce the number of parameters that needs to be learned while maintaining modeling accuracy, the concept of StochasticNets~\cite{stochasticnet} was introduced, where random graph theory~\cite{Erdos} is leveraged to form sparsely-connected deep neural networks via stochastic connectivity between neurons.  While StochasticNets have been shown to greatly reduce the number of parameters that needs to be learned, and thus effectively reduce the amount of training data, there are still many situations where there is insufficient training data in the domain of interest.

A recent field of research that attempts to tackle this issue is \textit{transfer learning}, where the goal is to transfer knowledge from other domains to the domain of interest.  Research have shown~\cite{Tlearning} the performance can be greatly improved via transfer learning without the need of extensive data labeling efforts for the training step.  Motivated by the potential of leveraging training data from another domain to boost modeling performance in the domain of interest, we study the efficacy of transfer learning in StochasticNets.

 \vspace{- 0.5 cm}
\section{Methodology}
\vspace{-0.4 cm}
In this paper, we are interested in studying efficacy of transfer learning in StochasticNets for addressing the challenge of small training data sizes in the domain of interest.  To this end, we apply transfer learning on deep convolutional StochasticNets in the following manner.  

First, a StochasticNet is trained based on the CIFAR-10 dataset \cite{CIFAR10}, which consists of 50,000 training images categorized into 10 classes which are not in the domain of interest.  Second, all the learned parameters from the convolutional layers from this trained StochasticNet are transferred to a new StochasticNet, one that is designed to perform classification in the domain of interest. Third, the parameters of the hidden layers of this new StochasticNet is trained using only 20\% of the STL-10 dataset~\cite{STL10}, which contains images categorized into 10 classes that are in the domain of interest.  For comparison purposes, a StochasticNet is trained directly using the same 20\% of the STL-10 dataset as the StochasticNet.
Figure~\ref{fig:Result} demonstrates the training and testing errors for both the baseline StochasticNet as well as the StochasticNet with transferred knowledge to be used as the reference baseline.

The deep convolutional StochasticNets used in this paper are realized based on the LeNet-5 deep convolutional neural network architecture~\cite{MNIST}, and consists of three convolutional layers with 32, 32, and 64 local receptive fields of size $5 \times 5$ for the first, second, and third convolutional layers, respectively, and one hidden layer of 64 neurons, with all neural connections being randomly realized based on probability distributions.  

The deep convolutional StochasticNets used in this paper were formed with 75\% neural connectivity via a Gaussian connectivity model, where the mean is at the center of the receptive field and the standard deviation is set to be a third of the receptive field size as with previous studies~\cite{stochasticnet}.

\section{Discussion}
\begin{figure}[tp]
\begin{center}
\includegraphics[scale = 0.35]{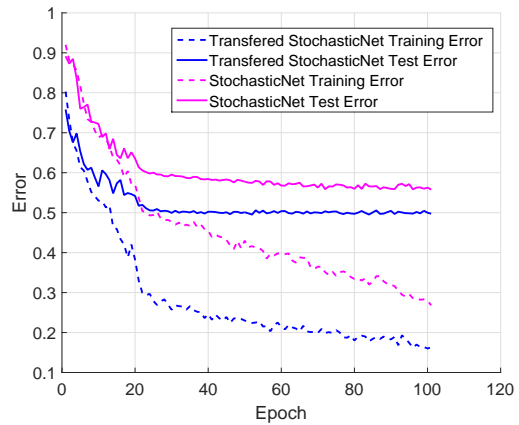}
\end{center}
\vspace{-0.5 cm}
\caption{Training and test error for the STL-10 dataset.  Blue lines represent the performance result of the StochasticNet with transferred knowledge, while the purple lines represent the performance result of the baseline StochasticNet. It can be observed that the StochasticNet with transferred knowledge achieves a significant improvement in performance ($\sim$ 7\%) when compared to the baseline StochasticNet.  }
\vspace{-0.5 cm}
\label{fig:Result}
\end{figure}

\vspace{-0.4 cm}
Figure~\ref{fig:Result} demonstrates the training and test errors for both the baseline StochasticNet as well as the StochasticNet with transferred knowledge.   It can be observed that the StochasticNet with transferred knowledge provides significant improvements in classification accuracy when compared to the baseline StochasticNet, with a decrease in test error of $\sim$ 7\%.  These very promising preliminary results demonstrate the efficacy of transfer learning on StochasticNets, and highly motivate future deeper exploration into this area for boosting the performance of StochasticNets in the situation where there is insufficient training data in the domain of interest.

\vspace{- 0.5 cm}
\section*{Acknowledgments}
\vspace{-0.4 cm}
This work was supported by the Natural Sciences and Engineering Research Council of Canada, Canada Research Chairs Program, and the Ontario Ministry of Research and Innovation. The authors also thank Nvidia for the GPU hardware used in this study through the Nvidia Hardware Grant Program.

\end{document}